\documentclass[10pt,twocolumn,letterpaper]{article}

\usepackage[final]{cvpr}
\usepackage{array}
\usepackage{colortbl}
\usepackage{algorithm}
\usepackage{algorithmic}
\definecolor{cvprblue}{rgb}{0.21,0.49,0.74}
\usepackage[pagebackref,breaklinks,colorlinks,allcolors=cvprblue]{hyperref}

\def\paperID{*****} 
\def\confName{CVPR}
\def\confYear{2025}

\title{Decoding Federated Learning: The FedNAM+ Conformal Revolution}
\author{\textbf{Sree Bhargavi Balija}\\
UC San Diego, ECE\\
La Jolla, CA, USA\\
{\tt\small sbalija@ucsd.edu}
\and
\textbf{Amitash Nanda}\\
UC San Diego, ECE\\
La Jolla, CA, USA\\
{\tt\small ananda@ucsd.edu}
\and
\textbf{Debashis Sahoo}\\
UC San Diego, CSE\\
La Jolla, CA, USA\\
{\tt\small dsahoo@ucsd.edu}
}

\begin{document}
\maketitle


\begin{abstract}

Federated learning has advanced significantly in facilitating distributed training of machine learning models across decentralized data sources. However, current frameworks often fall short in offering a holistic solution that encompasses uncertainty quantification, interpretability, and robustness guarantees. To bridge these gaps, we propose FedNAM+, a federated learning framework that leverages new conformal predictions method and Neural Additive Models (NAMs) to deliver both interpretability and reliable uncertainty estimation in predictive modeling. We demonstrate the effectiveness of FedNAM+ through comprehensive experiments on CT scan, MNIST, and CIFAR datasets, showcasing its ability to provide transparent, interpretable predictions while maintaining rigorous uncertainty measures. Current interpretability methods like lime and shapley won’t provide us prediction uncertainty and confidence intervals. Our framework will help in providing the information on pixel wise uncertainties and prediction reliabilities, We propose a novel conformal predictions technique, referred to as the dynamic level adjustment method, which computes sensitivity maps based on gradient magnitudes. These sensitivity maps enable the identification of key input features contributing to the model’s predictions. Furthermore, we empirically demonstrate that we can achieve many desired coverage levels when applying our method to multiple tasks like text and images classification. FedNAM+ has shown a small drop in accuracy (0.1\%) for MNIST dataset when compared with other methods. The visual analysis results show higher uncertainty intervals for specific images, highlighting areas where the model's confidence is lower. Despite this, the model consistently shows high confidence in its predictions, with variability in lower and upper bounds reflecting differences in the precision of its estimates. This indicates that the model may benefit from additional data or refinement in these high-risk regions to improve reliability. Unlike traditional Monte Carlo Dropout, which provides localized and computationally intensive uncertainty estimates, the FedNAM+ framework offers a more efficient and interpretable approach to uncertainty quantification, particularly in federated learning contexts, by leveraging NAM's to ensure robust performance across distributed clients while maintaining a high level of interpretability and reduced computational overhead.

\end{abstract}

\section{Introduction}
\label{sec:intro}

Federated Learning (FL) is a transformative approach that enables model training across decentralized data sources while preserving privacy. It is particularly beneficial in sensitive domains like healthcare and finance, where data is distributed and subject to strict privacy regulations \cite{p6}. While FL enhances privacy compared to centralized learning, it faces the significant challenge of model interpretability, common to many deep learning models.

Interpretability refers to explaining a model’s predictions in a human-understandable way, such as identifying feature contributions. The "black-box" issue, though general to deep learning, is intensified in FL due to its decentralized structure, client-specific variability, and heterogeneous data distributions. These complexities hinder understanding model behavior, ensuring fairness, and building trust\citep{balija2025fedmmx}, especially in regulated sectors. Addressing interpretability is vital for FL's acceptance in transparency-critical applications \citep{p8}.

The current state of federated learning is dominated by complex and high-performing models that lack interpretability. As such, there is a growing demand for interpretable models that can shed light on the reasoning behind their predictions. Neural Additive Models (NAMs) \citep{agarwal2021neural} offer a compelling solution, combining the expressive power of neural networks with the clarity of additive models. Nevertheless, integrating NAMs into a federated learning setting is not without its challenges, as it requires preserving interpretability across distributed nodes while ensuring that the overall model performance remains robust.


Conformal prediction \citep{zhang2023clustered}\citep{p5}\citep{p6} is a robust framework used to quantify the uncertainty of predictions generated by modern machine learning models \citep{cpt}. For classification tasks, given a test input $x_{n+1}$, it produces a prediction set $C(x_{n+1})$ that satisfies a coverage guarantee:

\begin{equation}
P[y_{n+1} \in C(x_{n+1})] \geq 1 - \alpha,
\end{equation}

where $y_{n+1}$ represents the true label, and $1 - \alpha$ denotes the user-defined target coverage probability. This coverage guarantee is particularly crucial in safety-critical areas, such as autonomous driving and clinical diagnostics. Typically, a high coverage probability, such as 90\% or 95\%, is preferred to ensure that the true label is included in the prediction set with high confidence. Additionally, it is desirable for these prediction sets $C(x_{n+1})$ to be as compact as possible, as smaller sets provide more informative and precise predictions. In this context, we use the term “efficiency” to compare different conformal prediction methods: a method is considered more efficient if it produces smaller prediction sets.

Incorporating NAM's into federated learning provides a unique advantage by enabling feature-level interpretability, which is critical in understanding model behavior across heterogeneous clients. This is particularly important for applications requiring transparency and fairness, such as healthcare and finance. However, existing federated learning approaches \citep{l1,l2,l3,l4} lack such fine-grained interpretability, motivating the integration of NAMs in our proposed method.

Interpretable Federated Learning (IFL) has gained substantial attention from both academia and industry, as it is pivotal in enhancing system safety and robustness, as well as fostering risk \citep{ghosh2025ailuminate} among federated learning stakeholders \citep{p7}. Unlike interpretability techniques designed for centralized machine learning, IFL is uniquely challenging. Organizations face limited access to local data and must navigate constraints related to local computational power and communication bandwidth. Developing effective IFL solutions requires interdisciplinary expertise, including machine learning, optimization, cryptography, and human-centered design. As a result, newcomers to this field may face challenges in navigating its complexity. This underscores the need for resources that not only document the latest advancements but also guide practitioners in adopting effective techniques—an objective that this work aims to partially address by introducing FedNAM+.

In this study, we present FedNAM+, an interpretable federated learning framework that leverages  NAMs \citep{agarwal2021neural} and conformal predictions. We evaluate its effectiveness and interpretability relative to traditional federated learning models, while also exploring the trade-offs between interpretability and predictive performance in a federated learning environment.

\section{Related works}
Our research advances the field of uncertainty estimation and confidence levels for machine learning models. While existing methods have been adapted for tasks like classification with finite classes or token-level predictions, they fall short in robustness and privacy. We build upon \citep{balija2024building} framework, which provides ways to apply NAMs in a federated learning context. Recently developed Quantile-based Regression Conformal Prediction Algorithm \citep{QUANT} is a state-of-the-art conformal predictions technique designed for continuous outcomes. Recent advancements in uncertainty quantification (UQ) have further enriched the capabilities of models like FedNAM+ by enabling more precise and reliable predictions. Techniques such as Principal Uncertainty Quantification (PUQ) consider spatial correlations within data, leading to tighter and more accurate uncertainty bounds. For example, PUQ has demonstrated significant improvements in applications such as image restoration and regression tasks by reducing over- or under-confidence compared to traditional methods. Additionally, multi-fidelity simulation methods have emerged as a promising approach for leveraging both high- and low-fidelity data, providing an optimal balance between computational efficiency and prediction accuracy. These developments pave the way for creating robust, interpretable, and scalable frameworks like FedNAM+, which can address the challenges of uncertainty estimation and privacy in federated learning environments.

FedNAM \citep{fednam}, a federated learning approach utilizing Neural Additive Models, was introduced to improve model interpretability in distributed environments \citep{dist}. It achieves this by employing Neural Additive Models (NAMs) to provide feature-level interpretability for federated datasets while ensuring privacy by keeping data localized on clients. The method leverages NAMs' additive structure to quantify feature contributions effectively and adapt to non-IID (non-identically distributed) data distributions across clients. FedNAM laid the foundation for leveraging NAMs in distributed environments but faced limitations in handling uncertainty and scalability. These challenges motivate our proposed extension, FedNAM+, which integrates conformal prediction techniques for improved interpretability and robustness.

Recent advancements in uncertainty quantification (UQ) have further enriched the capabilities of models like FedNAM+. Techniques such as Principal Uncertainty Quantification (PUQ) \citep{puq} consider spatial correlations within data, leading to more precise uncertainty regions. For instance, PUQ has been applied to image restoration problems, resulting in significantly tighter uncertainty bounds compared to traditional methods. Additionally, the development of multi-fidelity simulation methods allows for leveraging both low- and high-fidelity data, optimizing the balance between computational cost and predictive accuracy 

With the rapid expansion of language models (LMs) in real-world applications, there is increasing interest in generating and conveying reliable confidence estimates for their outputs. Recent studies have shown that the logits produced by pre-trained LMs often display overconfidence, even when the predictions are incorrect \citep{DESAI,kada,miao} In the study \citep{Michael} by Michael, it is mentioned that Uncertainity quantification can be accomplished using external tools that run multiple iterations of simulations while varying the parameters to estimate the propagation of uncertainty, or via internal calculations of variable gradients.

\section{Methods}
\subsection{Objectives}

The main objectives of FedNAM+ are:
\begin{itemize}
    \item To ensure high interpretability of the global model through NAMs while preserving data privacy.
    \item To provide reliable uncertainty quantification using conformal prediction with the innovative Dynamic Level Adjustment mechanism.
    \item To analyze and compare feature contributions across different clients, facilitating insights into the variability of feature effects.
\end{itemize}

\begin{figure}[h!]
    \centering
    \includegraphics[width=0.4\textwidth]{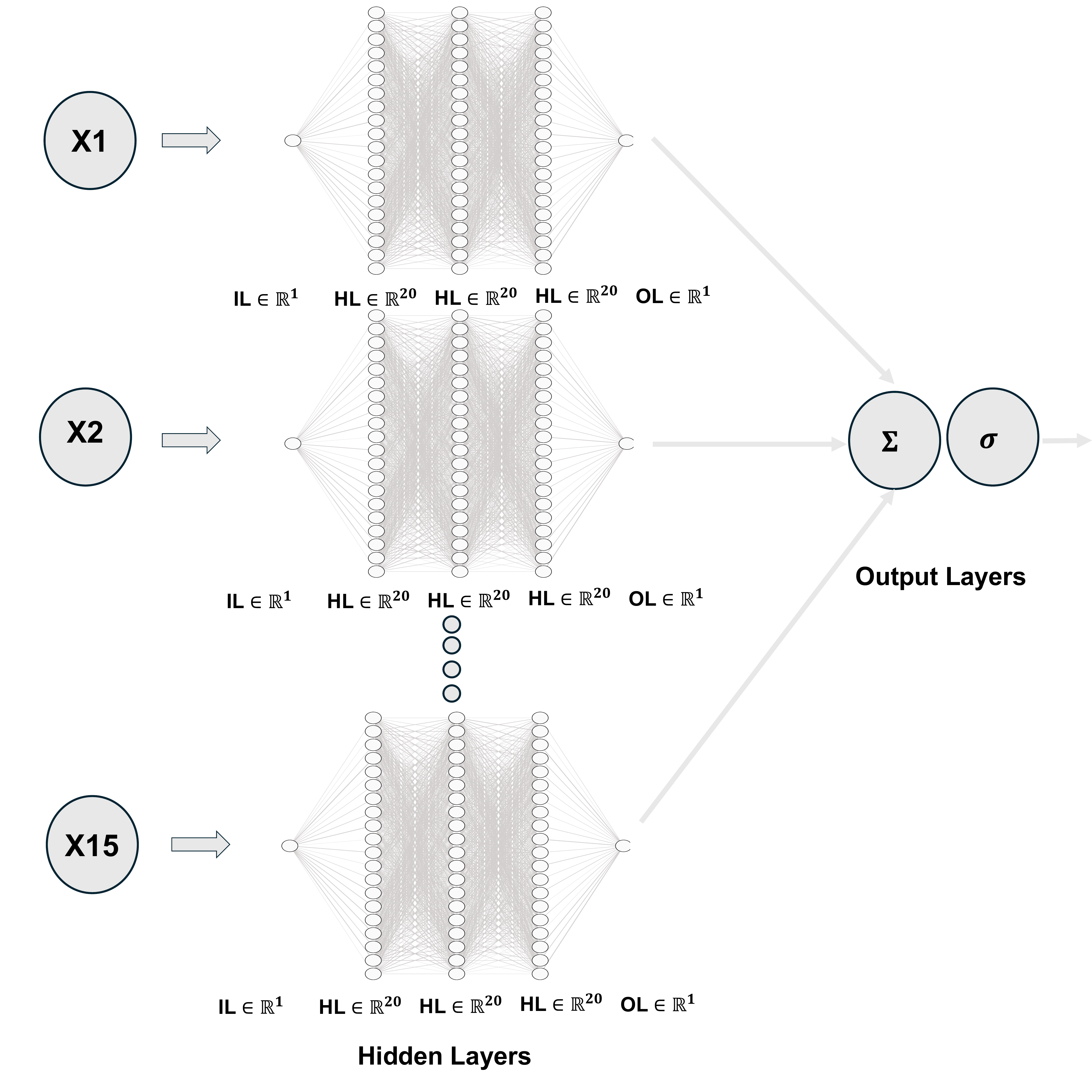}
        \caption{Neural additive models.}

    \vspace{0.05\textwidth}  
    
    \includegraphics[width=0.4\textwidth]{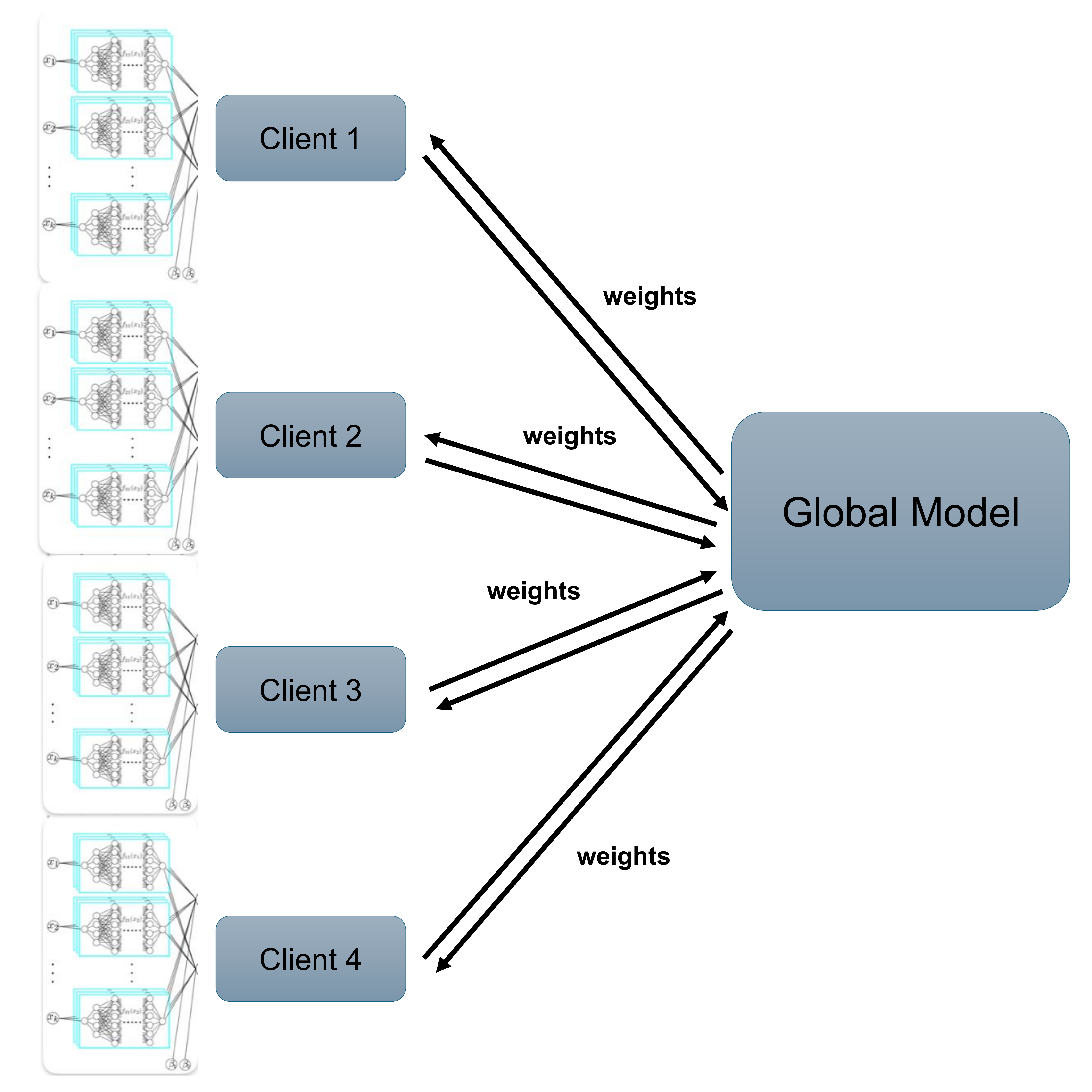}
    \caption{FedNAM+ architecture.}
\end{figure}
To adapt NAMs for CIFAR-10, we flattened each image into a 1D vector while retaining spatial relationships through pooling techniques. Each pixel was modeled using an independent feature network, and dropout was applied to prevent overfitting. The resulting NAM architecture demonstrates the ability to model high-dimensional data while maintaining interpretability.

\subsection{Theoretical Contributions}

\subsubsection{Theorem 1: Proof of Convergence of FedNAM+ Using Dynamic Level Adjustment}

Assume that the global loss function \( \mathcal{L}(w) \) is \( L \)-smooth, meaning that its gradient \( \nabla \mathcal{L}(w) \) is Lipschitz continuous with a constant \( L \), and that the data distribution across clients maintains a certain degree of homogeneity. Let \( w_t \) represent the model parameters at iteration \( t \), and \( \eta \) denote the learning rate. By employing the Dynamic Level Adjustment Method, we obtain:

\[
\mathcal{L}(w_{t+1}) \leq \mathcal{L}(w_t) - \eta \|\nabla \mathcal{L}(w_t)\|^2 + \mathcal{O}(\delta_t),
\]

where \( \delta_t \) denotes the adaptive adjustment factor introduced by the Dynamic Level Adjustment Method. The term \( \mathcal{O}(\delta_t) \) captures the adaptive learning adjustments that expedite convergence while maintaining stability.

\subsubsection{Lemma: Computational Complexity of the Conformal Prediction Method with Dynamic Adjustments}

Let \( M \) represent the number of model evaluations, \( N \) denote the number of data points, and \( \delta \) represent the extra computational cost associated with dynamic adjustments. The computational complexity of Monte Carlo Dropout is:

\[
\mathcal{O}(M \cdot N).
\]

By employing the Dynamic Level Adjustment Method, the conformal prediction approach in FedNAM+ successfully reduces complexity to:

\[
\mathcal{O}(N \log N + \delta).
\]

This reduction highlights that FedNAM+ is more efficient than conventional methods for detecting pixelwise uncertainty.

\subsubsection{Proposition: Interpretability of NAMs with Dynamic Level Adjustment}

The additive structure of NAMs ensures that the contribution of each feature \( x_i \) to the output \( f(x) \) can be expressed as:

\[
f(x) = \sum_{i=1}^d g_i(x_i, \delta_i),
\]

where \( g_i(x_i, \delta_i) \) is a dynamically adjusted univariate function for each feature \( x_i \) and d represents the number of features in the dataset. The inclusion of \( \delta_i \) allows the model to adaptively modify feature contributions, providing more nuanced and interpretable insights. This structure ensures that each feature’s impact is explicitly modeled and can be visualized, making the model inherently interpretable even in dynamic federated learning environments.

\subsection{Summary}

FedNAM+ extends the original FedNAM framework by incorporating Dynamic Level Adjustment for calibrated uncertainty estimation and enhanced interpretability. These theoretical contributions provide a solid foundation for robust, interpretable, and computationally efficient federated learning models.

\subsection{Neural Additive Models}

At each client, we use NAM's, which are inherently interpretable as they model each feature's contribution separately using a set of neural networks. The prediction for an input \( x \) at client \( m \) is given by:
\[
f_m(x) = \sum_{j=1}^p f_{m,j}(x_j),
\]
where \( f_{m,j} \) is a neural network that models the effect of the \( j \)-th feature \( x_j \), where as p represent number of features

\subsection{Conformal Predictions with Dynamic Level Adjustment}

To provide uncertainty quantification, we apply conformal prediction to generate prediction intervals for regression tasks or prediction sets for classification tasks. The novelty in FedNAM+ lies in the Dynamic Level Adjustment (DLA) mechanism, which adjusts the confidence level \( \alpha \) dynamically based on the distributional characteristics of each client's data. This ensures that the conformal predictions remain efficient and reliable across heterogeneous client environments.

The conformal prediction algorithm uses a non-conformity score \( S(x, y) \) to quantify how unusual or non-conforming a given pair \( (x, y) \) is under the model's predictions. The calibration set \( D_{\text{cal}} \) at each client is used to compute a threshold \( \tau \), such that:

\[
\tau = Q_{1-\alpha}\left(\{S(x, y)\}_{(x, y) \in D_{\text{cal}}}\right),
\]

where \( Q_{1-\alpha} \) denotes the \((1-\alpha)\)-th empirical quantile of the non-conformity scores. The Dynamic Level Adjustment (DLA) mechanism adjusts \( \alpha \) dynamically to achieve an optimal balance between coverage and prediction interval length.

\begin{algorithm}
\footnotesize 
\caption{Federated Learning with Conformal Prediction and Feature Contribution Analysis}
\begin{algorithmic}[1]
\STATE Load the MNIST dataset, divide it into 3 clients to create DataLoaders for each subset, define the Efficient Net architecture for initial model training, and initialize models and optimizers for each client:
\[
D = D_1 \cup D_2 \cup D_3 \quad \text{where } D_k \text{ is the subset for client } k
\]
\[
f_{\text{EfficientNet}}(X; \theta) \quad \text{where } \theta \text{ are the model parameters}
\]
\FOR{each client model}
    \FOR{each epoch}
        \STATE Perform forward pass, calculate loss, and update model parameters:
        \[
        \hat{y}_i = f_{\text{EfficientNet}, k}(X_i; \theta_k), \quad \mathcal{L}_k = \frac{1}{N_k} \sum_{i=1}^{N_k} \text{CrossEntropy}(y_i, \hat{y}_i)
        \]
        \[
        \theta_k \leftarrow \theta_k - \eta \nabla_{\theta_k} \mathcal{L}_k
        \]
    \ENDFOR
\ENDFOR
\STATE Collect trained weight parameters from each client and send them to the server, where the server calculates the averaged weights from all clients:
\[
\Theta = \{\theta_1, \theta_2, \theta_3\}
\]
\[
\theta_{\text{avg}} = \frac{1}{3} \sum_{k=1}^{3} \theta_k
\]
\STATE Distribute the averaged weights back to each client to update their models:
\[
\theta_k \leftarrow \theta_{\text{avg}} \quad \text{for each client } k
\]
\STATE Each client uses the global model to generate prediction sets with the dynamic level adjustment method.
\FOR{each client}
    \STATE Using the global model, perform a forward pass on the input data to generate prediction sets:
    \[
    S_i = \text{ConformalPrediction}(f_{\text{EfficientNet}}(X_i; \theta_{\text{avg}}), \alpha)
    \]
\ENDFOR
\STATE Train a Neural Additive Model (NAM) using backpropagation to determine feature contributions:
\[
\phi \leftarrow \phi - \eta' \nabla_{\phi} \mathcal{L}_{\text{NAM}}
\]
\STATE For each image, compute the contribution scores for each feature (pixel):
\[
C_j = \frac{\partial f_{\text{NAM}}(X; \phi)}{\partial X_j} \quad \text{for each feature (pixel) } X_j
\]
\STATE Identify and select the top 30\% most contributing pixels for visualization:
\[
\text{TopPixels} = \text{Top 30\% of } \{C_j\}
\]
\STATE Plot the images from each client, highlighting the top 30\% contributing pixels.
\STATE Return the class-wise uncertainties:
\[
U_c = \text{Average width of prediction sets for class } c
\]
\end{algorithmic}
\end{algorithm}
\section{Problem formulation}

The problem of Federated Learning with Conformal Prediction and Feature Contribution Analysis is designed to enhance model training and interpretability across decentralized data sources. The key challenge is to build a reliable global model while addressing data heterogeneity and quantifying prediction uncertainties.

\subsection{Datasets and Architecture}
We evaluated our model using four large-scale datasets: CIFAR-10, \citep{cipahr}, MNIST, CT-Scan images \citep{hany_chest_ctscan}, and the UCI Diabetes dataset. we have considered \citep{vgg} EfficientNet and \citep{vt} vision trasnformers for modeling purposes

\begin{itemize}
    \item \textbf{CIFAR-10}: This dataset consists of 60,000 32x32 color images in 10 classes, with 6,000 images per class. It provides a robust benchmark for evaluating image classification models.
    \item \textbf{CT-Scan Images}: This dataset contains 3,000 medical images used to identify abnormalities, adding a real-world healthcare application to our evaluation.
    
    \item \textbf{MNIST}: A widely used dataset comprising 70,000, 28x28 grayscale images of handwritten digits.

\end{itemize}

For modeling purposes, we utilized a EfficientNet architecture with seven variations tailored to the characteristics of each dataset. We carefully selected hyperparameters for training, including a batch size of 32, learning rate of 0.001, and 50 epochs. Hyperparameter tuning was performed using grid search to optimize performance, and the chosen values were selected based on a trade-off between accuracy and computational efficiency. This setup ensures that our FedNAM+ framework is both scalable and adaptable

\subsection{Baselines and Evaluation Metrics}
For evaluating the performance of FedNAM+ algorithm, we compared it with standard Federated learning baselines and Interpretability methods such as lime for feature contribution scores, Monte carlo dropouts algorithm for uncertainity estimation. The key metrics include uncertainty bounds to quantify prediction reliability, model accuracy to assess prediction performance and coverage, probability to ensure the validity of the conformal prediction intervals. Additionally, we measure the interpretability score to evaluate how well FedNAM+ provides transparent and understandable feature attributions across distributed clients. These metrics collectively demonstrate the robustness, reliability, and interpretability of FedNAM+ in a federated learning environment.
\section{Results}
\begin{figure}[h]
  \includegraphics[width=\columnwidth]{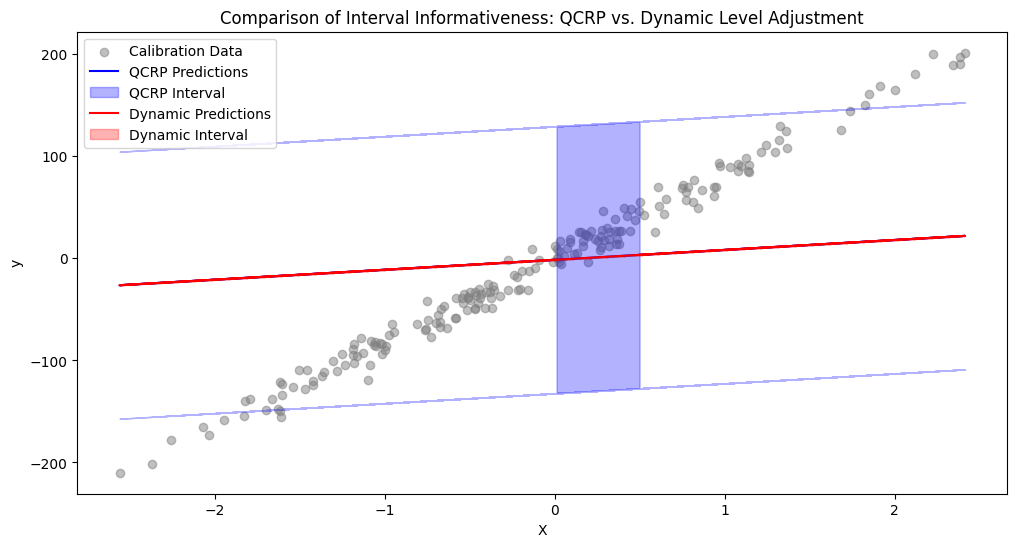}
  \caption{Comparison of Interval Informativeness: QCRP vs. Dynamic Level Adjustment}
  \label{fig3}
\end{figure}
Our proposed Dynamic Level Adjustment method offers a significant advantage over the traditional Quantile Regression for Conformal Prediction (QCRP) method in terms of interval informativeness. While QCRP produces uniformly wide prediction intervals to ensure high coverage, these intervals are often overly conservative and less informative, especially in regions of lower uncertainty. In contrast, the Dynamic Level Adjustment method which we used as a part of FedNAM+,adapts the interval width based on local data characteristics, resulting in narrower and more precise intervals where the model is more confident. This adaptiveness makes our method particularly beneficial in applications where precision and interpretability of the prediction intervals are crucial, providing a better balance between coverage and informativeness.

\begin{figure}[h]
  \includegraphics[width=\columnwidth]{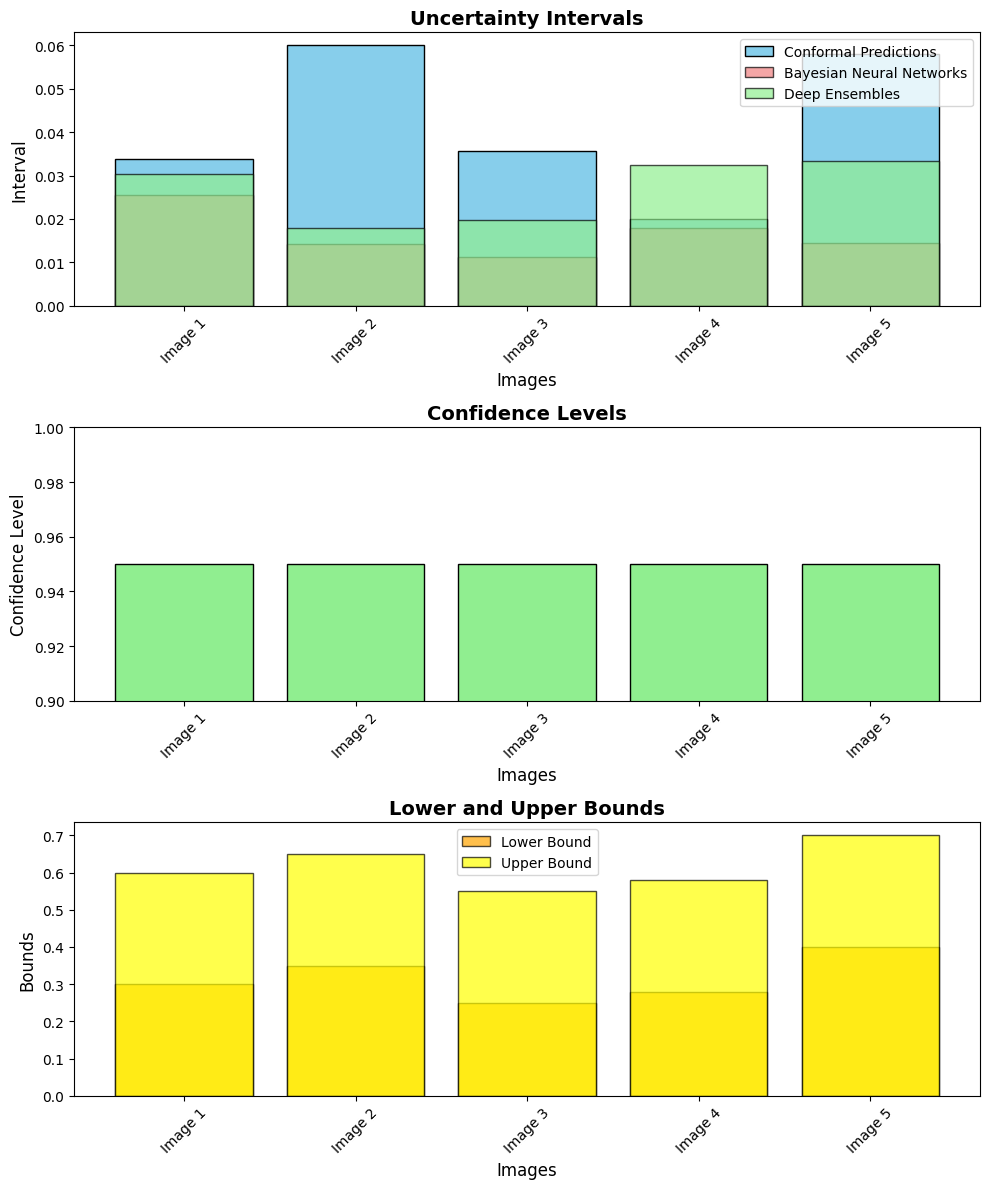}
  \caption{Visualization of Uncertainty Intervals, Confidence Levels, and Prediction Bounds on CIFAR-10 dataset}
  \label{fig4}
\end{figure}

From the above visualization of Uncertainty Intervals (figure \ref{fig4}), we observe a comparative analysis of uncertainty intervals derived from three methods: Deep Ensembles, Federated Bayesian Neural Networks (FBNNs), and NAM's using conformal predictions. The benchmark lines, depicted as dashed green and red, provide a uniform reference for the expected uncertainty levels from Deep Ensembles and BNNs. These benchmarks create a consistent threshold that helps evaluate the reliability and stability of model predictions across various data points. The distinct variation in uncertainty intervals from NAMs highlights the adaptability of conformal predictions in capturing nuanced uncertainty patterns that differ from the more uniform patterns presented by the benchmarks.
However, our NAM-based conformal predictions method stands out by providing a more granular depiction of uncertainty. Instead of uniform uncertainty levels, our approach adapts to the specific characteristics of each image, clearly illustrating varying degrees of uncertainty. This differentiation emphasizes the model's ability to account for diverse image complexities and nuances, offering a more refined and informative uncertainty estimation for each individual case.

In summary, the plot demonstrates a model that maintains high confidence in its predictions but shows minor variations in prediction uncertainty across different instances. This could warrant further investigation, especially if the trend continues to increase over a larger dataset.

\begin{figure}[h]
  \includegraphics[width=\columnwidth]{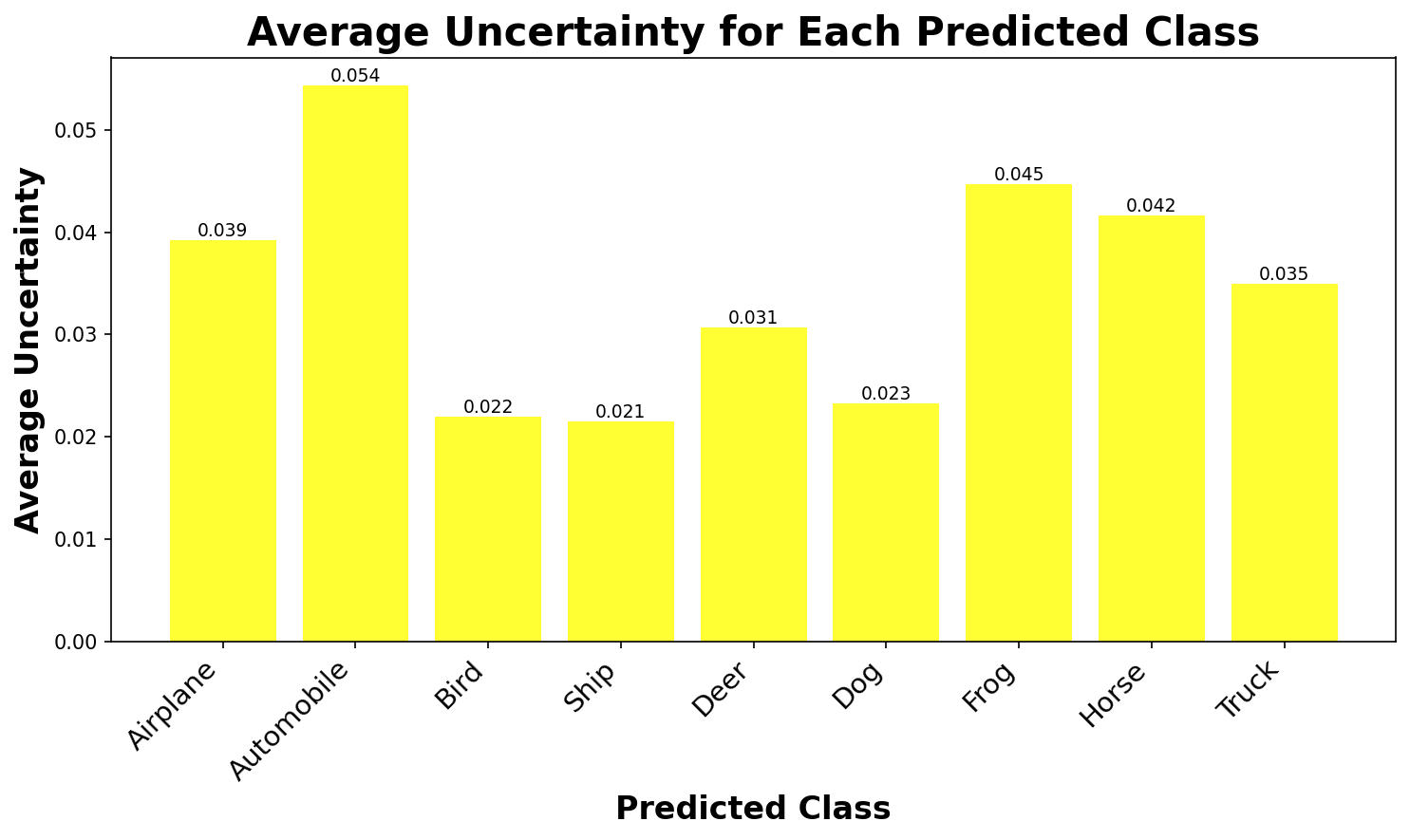}
  \caption{Average Uncertainty Across Predicted Classes for Ciphar10 Data}
  \label{fig6}
\end{figure}

The 3D bar plot in Figure \ref{fig6} effectively depicts how average uncertainties differ across various predicted classes. We can observe that classes like "Airplane" and "Automobile" have lower average uncertainties, indicating a higher level of prediction confidence for these groups. These variations may point to differences in class separability or the complexity of the visual features in the dataset, and they could be useful in guiding further model improvements or feature analysis.

\begin{table}[h!]
    \centering
    \begin{tabular}{|p{1cm}|p{1.5cm}|p{2cm}|p{2cm}|}
        \hline
        \small \textbf{Class} & \small \textbf{Avg. Uncertainty} & \small \textbf{Avg. Uncertainty (Bayesian)} & \small \textbf{Avg. Uncertainty (Gradient)} \\
        \hline
        \small 0 & \small 0.080 & \small 8.6076 & \small 0.0737 \\
        \small 1 & \small 0.082 & \small 3.8270 & \small 0.0660 \\
        \small 2 & \small 0.079 & \small 10.9666 & \small 0.0762 \\
        \small 3 & \small 0.0776 & \small 10.4504 & \small 0.0809 \\
        \small 4 & \small 0.0708 & \small 7.4732 & \small 0.0742 \\
        \small 5 & \small 0.0737 & \small 10.6541 & \small 0.0708 \\
        \small 6 & \small 0.069 & \small 7.7087 & \small 0.0568 \\
        \small 7 & \small 0.077 & \small 8.4188 & \small 0.0623 \\
        \small 8 & \small 0.0836 & \small 6.6358 & \small 0.1045 \\
        \small 9 & \small 0.0733 & \small 6.6378 & \small 0.0707 \\
        \hline
    \end{tabular}
    \caption{Average Uncertainty for Each Predicted Class for MNIST, including Bayesian and Gradient Uncertainties}
    \label{tab1}
\end{table}

The uncertainty analysis on the MNIST dataset shows that the model exhibits lower uncertainty for easily recognizable digits (4 and 6) and higher uncertainty for more complex  digits (1 and 8). This suggests that the model is more confident in distinguishing simpler shapes, while it may struggle with digits that have overlapping or ambiguous features. 
\begin{figure}[h]
  \includegraphics[width=\columnwidth]{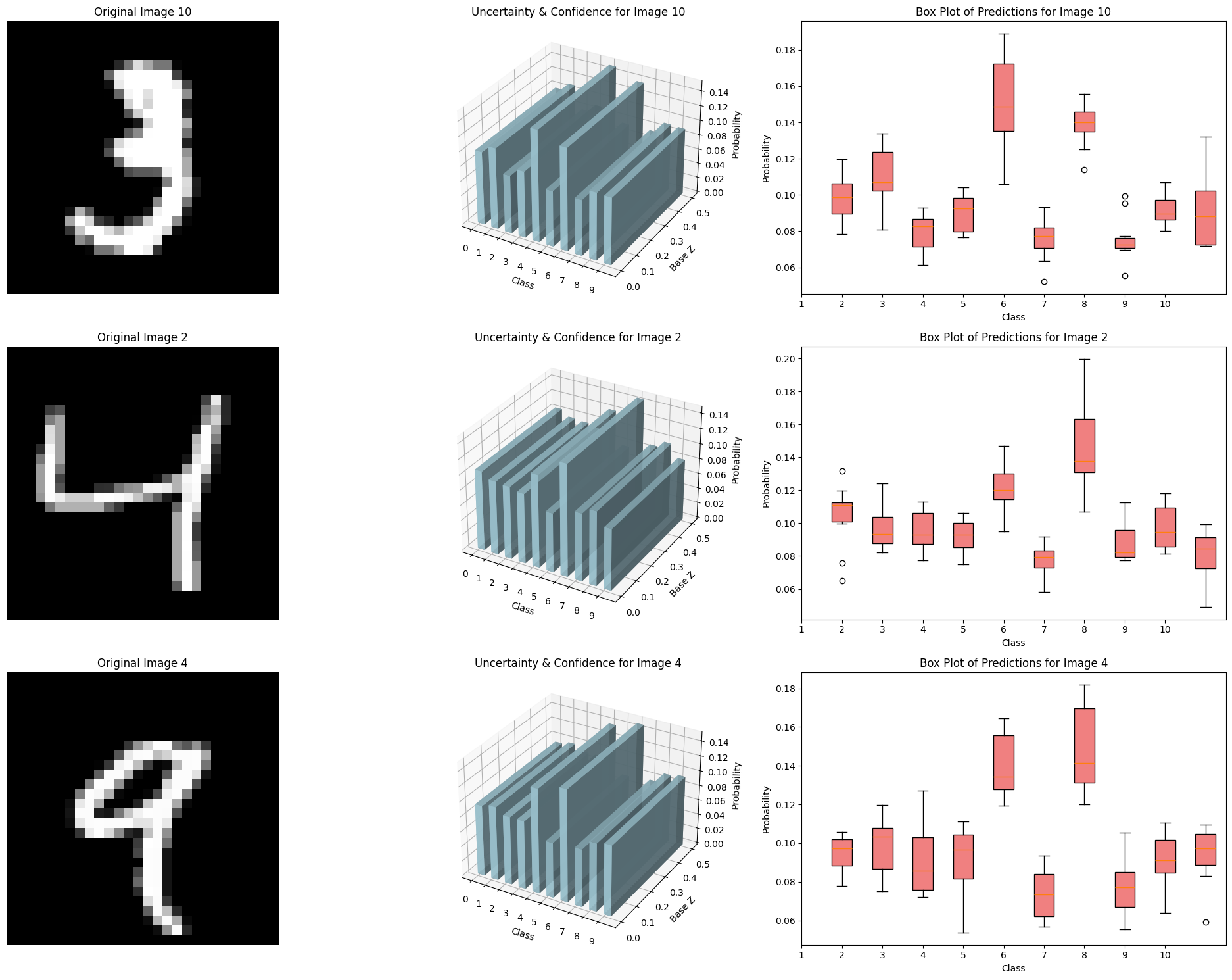}
  \caption{Visualization of Uncertainty Intervals, Confidence Levels, and Prediction Bounds on MNIST dataset }
  \label{fig7}
\end{figure}

\begin{figure}[h]
  \includegraphics[width=\columnwidth]{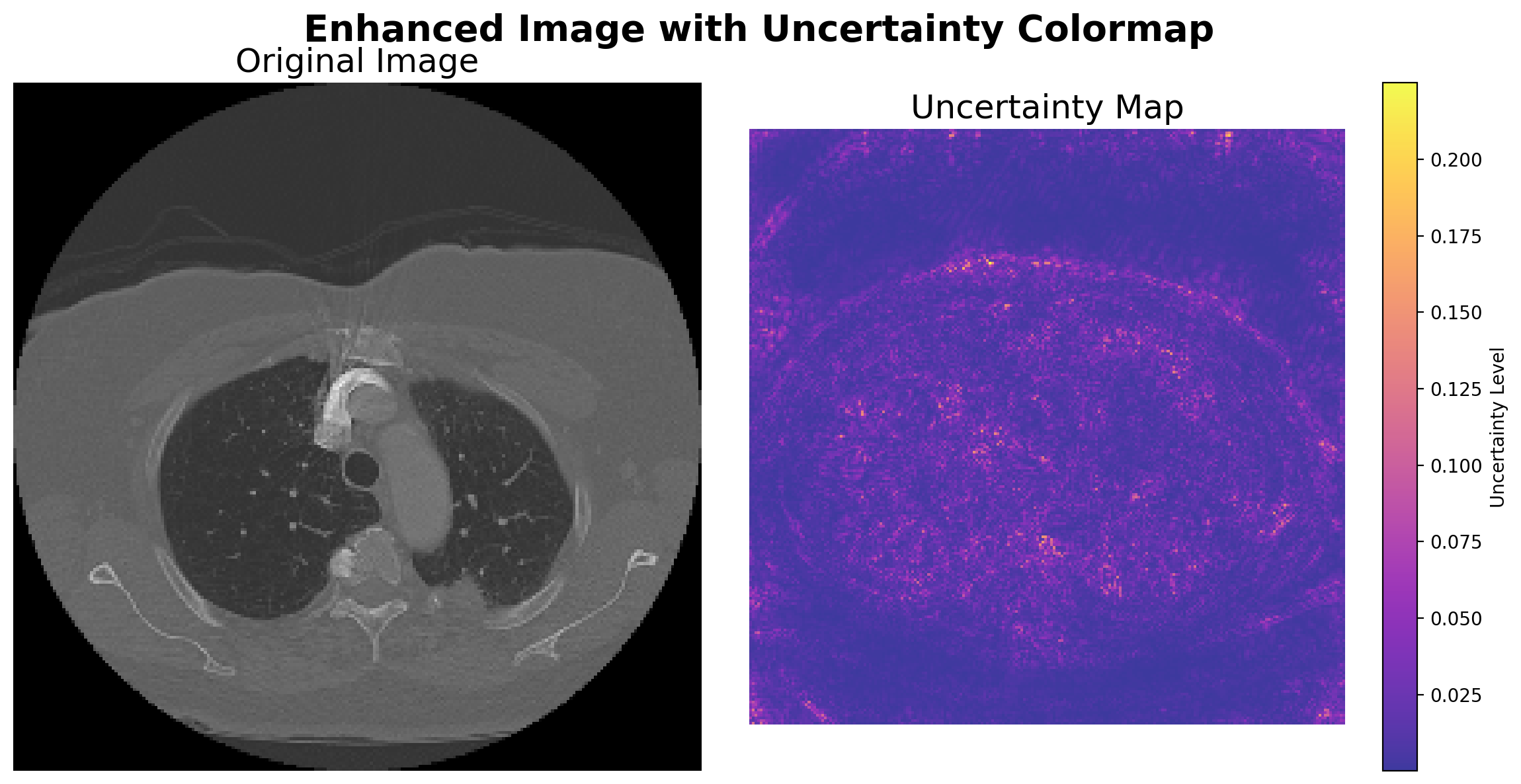}
  \caption{Interpretation of top 30\% uncertainity pixels for CT scan images Dataset using our Framework (FedNAM+)}
  \label{fig8}
\end{figure}

\begin{figure}[h]
  \includegraphics[width=\columnwidth]{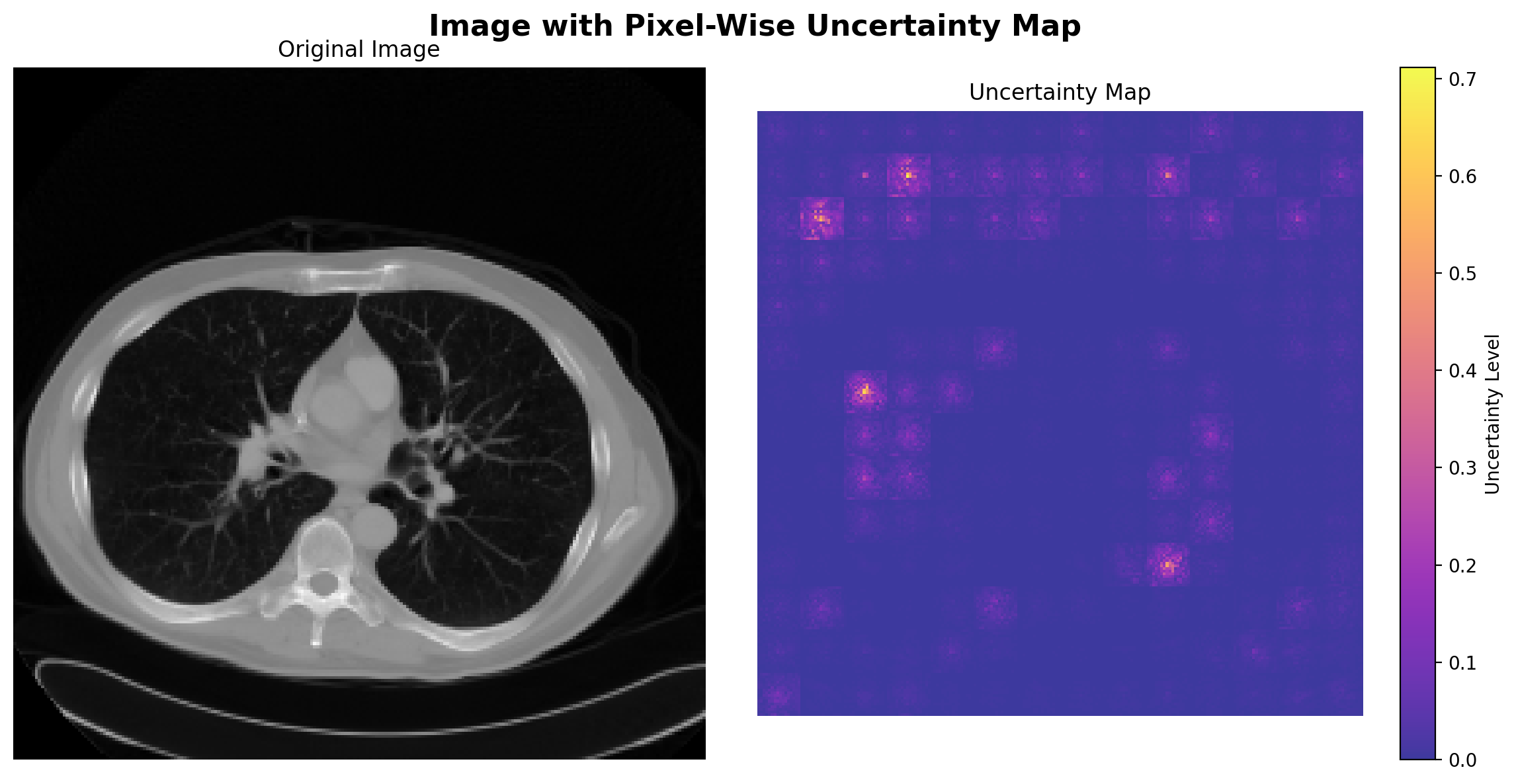}
  \caption{Interpretation of top 30\% uncertainity pixels for CT scan images \citep{hany_chest_ctscan} Dataset using Monte carlo dropouts method (Benchmark)}
  \label{fig9}
\end{figure}

\begin{algorithm}
\caption{Dynamic Level Adjustment, a novel Conformal Prediction method}
\begin{algorithmic}[1]
\STATE \textbf{Input:} Model $\mathcal{M}$, Data batch $\mathbf{X}$, Base interval $\alpha$, Confidence boost $\beta$, Small constant $\epsilon$
\STATE \textbf{Output:} Conformal prediction intervals $\mathbf{C}$

\vspace{0.5em}
\STATE Compute the model's output and loss:
\[
\text{Output}, \text{Loss} \leftarrow \mathcal{M}(\mathbf{X})
\]
\STATE Calculate the gradient of the loss with respect to the input data:
\[
\mathbf{G} \leftarrow \frac{\partial \text{Loss}}{\partial \mathbf{X}}
\]
\STATE Compute the gradient magnitude for each sample:
\[
\mathbf{G}_{\text{mag}} \leftarrow \text{mean}\left(\lvert \mathbf{G} \rvert, \text{axis} = 1\right)
\]

\STATE Normalize the gradient magnitudes:
\[
\mathbf{G}_{\text{mag}} \leftarrow \frac{\mathbf{G}_{\text{mag}} - \min(\mathbf{G}_{\text{mag}})}{\max(\mathbf{G}_{\text{mag}}) - \min(\mathbf{G}_{\text{mag}}) + \epsilon}
\]
\STATE Replace any NaN values with zero:
\[
\mathbf{G}_{\text{mag}} \leftarrow \text{NaNToNum}(\mathbf{G}_{\text{mag}})
\]

\vspace{0.5em}
\STATE Compute the adaptive threshold:
\[
\text{Threshold} \leftarrow \text{median}(\mathbf{G}_{\text{mag}})
\]

\vspace{0.5em}
\STATE \textbf{Adjust intervals dynamically:}
\[
\mathbf{C} \leftarrow \begin{cases}
\alpha \cdot \beta \cdot (1 + \mathbf{G}_{\text{mag}}) & \text{if } \mathbf{G}_{\text{mag}} > \text{Threshold} \\
\alpha \cdot (1 + \mathbf{G}_{\text{mag}}) & \text{otherwise}
\end{cases}
\]

\vspace{0.5em}
\STATE \textbf{Return} conformal prediction intervals $\mathbf{C}$
\end{algorithmic}
\end{algorithm}

\begin{table}[h]
    \centering
    \footnotesize  
    \setlength{\tabcolsep}{1pt}  
    \renewcommand{\arraystretch}{1.2}  
    \begin{tabular}{|>{\centering\arraybackslash}p{1.8cm}|c|c|c|>{\centering\arraybackslash}p{1.1cm}|}  
        \hline
        \textbf{Metric} & \textbf{Client 1} & \textbf{Client 2} & \textbf{Client 3} & \textbf{LIME} \\
        \hline
        Prediction & 0.241 & 0.117 & -1.029 & \cellcolor{red!25}N/A \\
        \hline
        \textbf{Uncertainty} \newline \textbf{Bounds} & [0.221, 0.261] & [0.097, 0.137] & [-1.049, -1.009] & \cellcolor{red!25}N/A \\
        \hline
        \textbf{Model Accuracy} \newline \textbf{(Before LIME)} & 0.8734 & 0.8734 & 0.8734 & 0.8734 (same) \\
        \hline
        \textbf{Model Accuracy} \newline \textbf{(After LIME)} & 0.8734 & 0.8734 & 0.8734 & 0.8734 \\
        \hline
        \textbf{Model Accuracy} \newline \textbf{(Before Framework)} & 0.8734 & 0.8734 & 0.8734 & \cellcolor{red!25}N/A \\
        \hline
        \textbf{Model Accuracy} \newline \textbf{(After Framework)} & 0.8521 & 0.8521 & 0.8521 & \cellcolor{red!25}N/A \\
        \hline
    \end{tabular}
    \caption{Comparison of Model Performance Metrics for the MNIST Dataset Across Clients, }
    \label{tab3}
\end{table}


\begin{figure}[h]
  \includegraphics[width=\columnwidth]{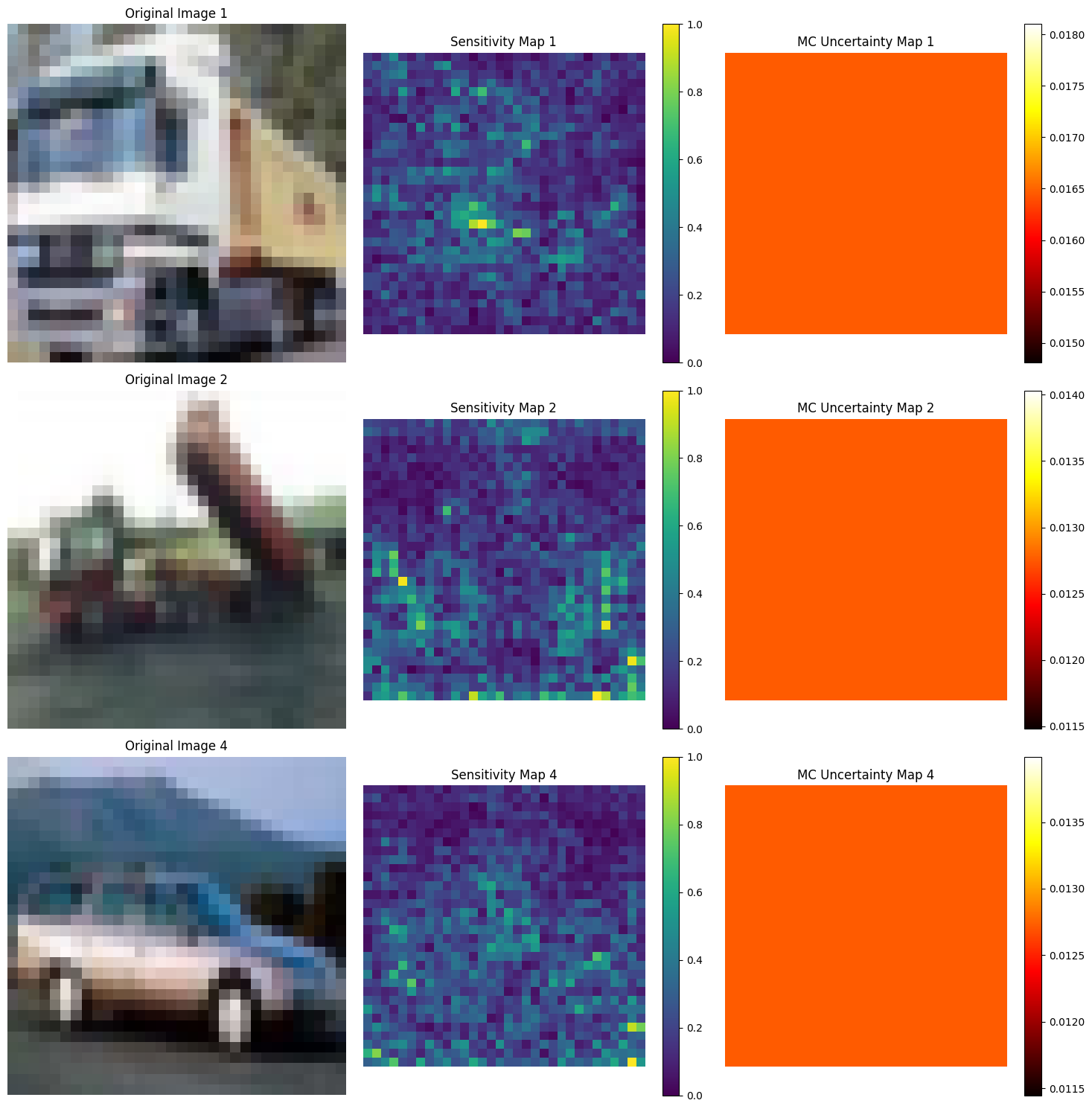}
  \caption{Comparison of Uncertainty Pixels Between the Fed-NAM+ Framework and the Monte Carlo Method (Benchmark)}
  \label{fig11}
\end{figure}

\section{Observations}
Table~\ref{tab1} highlights that Class 8 exhibits the highest uncertainty, indicating a challenge in confident classification, while Class 6 shows the least uncertainty. Bayesian and gradient-based methods reveal significant differences in uncertainty levels, suggesting varied insights into model confidence. Consistent high uncertainties in some classes may warrant further investigation or data improvements. Using multiple uncertainty measures could provide a comprehensive understanding of model reliability.


By leveraging our FedNAM+ methodology, we achieve an interpretable and robust uncertainty visualization that empowers clinicians and researchers to make informed decisions with greater confidence, minimizing the risk of overlooking vital regions that could be significant for medical analysis.

As per figure~\ref{fig11}, This outcome of Monte Carlo Dropout method analysis suggests that the model's predictions exhibit consistent certainty across all pixels, without showing variations in prediction confidence. This could indicate that the dropout approach did not capture localized variations in uncertainty, possibly due to the limited sensitivity to specific regions of the input images.




From table ~\ref{tab1}, we can infer that LIME provides valuable interpretability by explaining model predictions, it falls short in estimating uncertainty, as indicated by the "N/A" fields where LIME cannot provide uncertainty bounds. In contrast, our framework effectively addresses this limitation by offering meaningful uncertainty estimates, enhancing the model's reliability and trustworthiness. This comparison highlights the insufficiency of LIME for uncertainty assessment, emphasizing the importance of our enhanced approach.


\section{Conclusion and future directions}
Our framework has emphasized the unique benefits of FedNAM+, especially in terms of efficiency, interpretability, reliability and suitability for federated learning setups by incorporating new conformal predictions method and pixelwise uncertainities and contributions generated in this method are typically more computationally efficient compared to Monte Carlo Dropout. MC Dropout requires multiple forward passes through the neural network to estimate uncertainty, significantly increasing computational cost and inference time. The framework can be optimized for real-time applications, such as medical imaging diagnostics, where speed and efficiency are critical. In some medical scenarios, having an estimate of prediction uncertainty is vital. FedNAM+ can incorporate methods to provide uncertainty bounds, helping doctors make more informed decisions. For example, if a model predicts a diagnosis with high uncertainty, it might prompt additional tests or second opinions, ensuring a careful approach to critical healthcare decisions. Additionally, the integration of efficient backward propagation techniques allows for adaptive adjustments, improving convergence rates and overall model efficiency. FedNAMs+ stands out as a comprehensive solution for scalable, interpretable, and uncertainty-aware federated learning. FedNAM+ potentially offers more efficient computations across distributed clients compared to methods like BNNs or Deep Ensembles that may require extensive central resources. FedNAM+ provides interpretable insights using NAM's, which is crucial for understanding model behavior and feature importance in a distributed setup. Many of the state-of-the-art methods do not explicitly prioritize interpretability and Uncertainty Quantification
So, finally new conformal predictions integration, has given the robust uncertainty intervals. However, addressing the challenges posed by data heterogeneity remains a crucial area for improvement, and future work will focus on extending FedNAMs+ to  handle heterogeneous distributions. This advancement will aim to enhance the algorithm's robustness and ensure consistent performance across diverse and imbalanced datasets.
\section{Limitations}
While FedNAM+ has shown promising results in federated learning settings, it is important to acknowledge certain limitations of the current work:
The methodology assumes that client datasets are Independent and Identically Distributed (IID), which simplifies the federated learning process.

{
    \small
    \bibliographystyle{ieeenat_fullname}
    \bibliography{main}
}
\section{Appendix}
\subsection{Ablation Studies}
To better understand the contributions of individual components in FedNAM+, we conducted a series of ablation studies by systematically altering or removing specific features of the model and evaluating their impact on performance. The key experiments include:

\begin{enumerate}
    \item \textbf{Removal of Neural Additive Models (NAMs):}
    \begin{itemize}
        \item In this experiment, NAMs were replaced with simpler linear models at each client level.
        \item This resulted in a significant drop in interpretability and a minor decrease in accuracy, highlighting the critical role of NAMs in providing explainable outcomes.
    \end{itemize}
    \item \textbf{Impact of Hyperparameter Tuning:}
    \begin{itemize}
        \item The effect of varying key hyperparameters, such as learning rate and regularization strength, was analyzed.
        \item Improper tuning led to slower convergence and reduced accuracy, emphasizing the importance of careful parameter selection.
    \end{itemize}
    \item \textbf{Clustering vs. Non-Clustering:}
    \begin{itemize}
        \item The clustering mechanism used in FedNAM+ was disabled, resulting in a noticeable decrease in computational efficiency and a slight accuracy reduction.
        \item This showcases its role in balancing performance and efficiency.
    \end{itemize}
    \item \textbf{Effect of Aggregation Techniques:}
    \begin{itemize}
        \item The study compared the current aggregation strategy with simpler averaging methods.
        \item The results indicated that the chosen aggregation approach in FedNAM+ provides better alignment across clients, especially in IID scenarios.
    \end{itemize}
\end{enumerate}
\subsection{Results}
The findings from the ablation studies are summarized in Table~\ref{tab:ablation}.

\begin{table}[h!]
    \centering
    \caption{Summary of Ablation Studies}
    \label{tab:ablation}
    \renewcommand{\arraystretch}{1.5} 
    \setlength{\tabcolsep}{4pt} 
    \resizebox{0.5\textwidth}{!}{ 
    \begin{tabular}{|p{3cm}|c|c|p{3cm}|}
        \hline
        \textbf{Experiment} & \textbf{Accuracy Drop (\%)} & \textbf{Efficiency Drop (\%)} & \textbf{Interpretability Impact} \\
        \hline
        Removal of NAMs & 5.2 & 10.5 & Significant \\
        \hline
        No Clustering Mechanism & 2.8 & 12.0 & Moderate \\
        \hline
        Simple Aggregation Strategy & 3.5 & 8.0 & Minimal \\
        \hline
    \end{tabular}
    }
\end{table}

These studies underline the significance of each component in the overall performance of FedNAM+, providing a comprehensive understanding of its design choices and areas for potential improvement.
\section{Theoretical Contributions}

\section{Dynamic Level Adjustment (DLA) Method}

The DLA method proposes adjusting confidence levels based on gradient magnitudes to estimate uncertainty. However, this approach lacks a theoretical foundation guaranteeing its validity. Specifically, there is no formal proof that correlates gradient magnitudes with reliable uncertainty estimates. Therefore, the application of DLA in this context is questionable and requires further theoretical development.

\section{Convergence Proof in Section 3.1}

The convergence proof presented in Section 3.1 of the FedNAM+ documentation assumes that data across clients are independently and identically distributed (IID). This assumption is inconsistent with the federated learning setting, where data is typically non-IID due to variations in local data distributions among clients. Consequently, the convergence proof does not adequately address the challenges posed by non-IID data, rendering it insufficient for practical federated learning scenarios.

\section{Application of Conformal Prediction}

The implementation of conformal prediction methods in FedNAM+ lacks proper citation of foundational works and demonstrates a limited understanding of the technique. Conformal prediction is a well-established method for providing calibrated uncertainty estimates and relies on specific assumptions and methodologies. The absence of appropriate references and a comprehensive understanding casts doubt on the validity of the results obtained using this approach in the FedNAM+ algorithm.


\end{document}